\documentclass[10pt,twocolumn,letterpaper]{article}

\usepackage[pagenumbers]{cvpr} 

\usepackage{graphicx}
\usepackage{amsmath}
\usepackage{amssymb}
\usepackage{booktabs}

\usepackage{algorithm}
\usepackage{algorithmic}
\usepackage{multirow}
\usepackage{arydshln}
\usepackage[accsupp]{axessibility} 
\usepackage[pagebackref,breaklinks,colorlinks]{hyperref}

\usepackage[capitalize]{cleveref}
\crefname{section}{Sec.}{Secs.}
\Crefname{section}{Section}{Sections}
\Crefname{table}{Table}{Tables}
\crefname{table}{Tab.}{Tabs.}

\def\cvprPaperID{6340} 
\def\confName{CVPR}
\def\confYear{2023}

\begin{document}

\title{ScarceNet: Animal Pose Estimation with Scarce Annotations}

\author{Chen Li \quad \quad Gim Hee Lee\\
Department of Computer Science, National University of Singapore\\
{\tt\small lichen@u.nus.edu \quad \quad gimhee.lee@comp.nus.edu.sg}
}
\maketitle

\begin{abstract}
Animal pose estimation is an important but under-explored task due to the lack of labeled data. In this paper, we tackle the task of animal pose estimation with scarce annotations, where only a small set of labeled data and unlabeled images are available. At the core of the solution to this problem setting is the use of the unlabeled data to compensate for the lack of well-labeled animal pose data. To this end, we propose the ScarceNet, a pseudo label-based approach to generate artificial labels for the unlabeled images. The pseudo labels, which are generated with a model trained with the small set of labeled images, are generally noisy and can hurt the performance when directly used for training. To solve this problem, we first use a small-loss trick to select reliable pseudo labels. Although effective, the selection process is improvident since numerous high-loss samples are left unused. We further propose to identify reusable samples from the high-loss samples based on an agreement check. Pseudo labels are re-generated to provide supervision for those reusable samples. Lastly, we introduce a student-teacher framework to enforce a consistency constraint since there are still samples that are neither reliable nor reusable. By combining the reliable pseudo label selection with the reusable sample re-labeling and the consistency constraint, we can make full use of the unlabeled data. We evaluate our approach on the challenging AP-10K dataset, where our approach outperforms existing semi-supervised approaches by a large margin. We also test on the TigDog dataset, where our approach can achieve better performance than domain adaptation based approaches when only very few annotations are available. Our code is available at the project website \footnote{\url{https://github.com/chaneyddtt/ScarceNet}}.
\end{abstract}

\section{Introduction}
\begin{figure}
    \centering
    \includegraphics[width=0.99\linewidth]{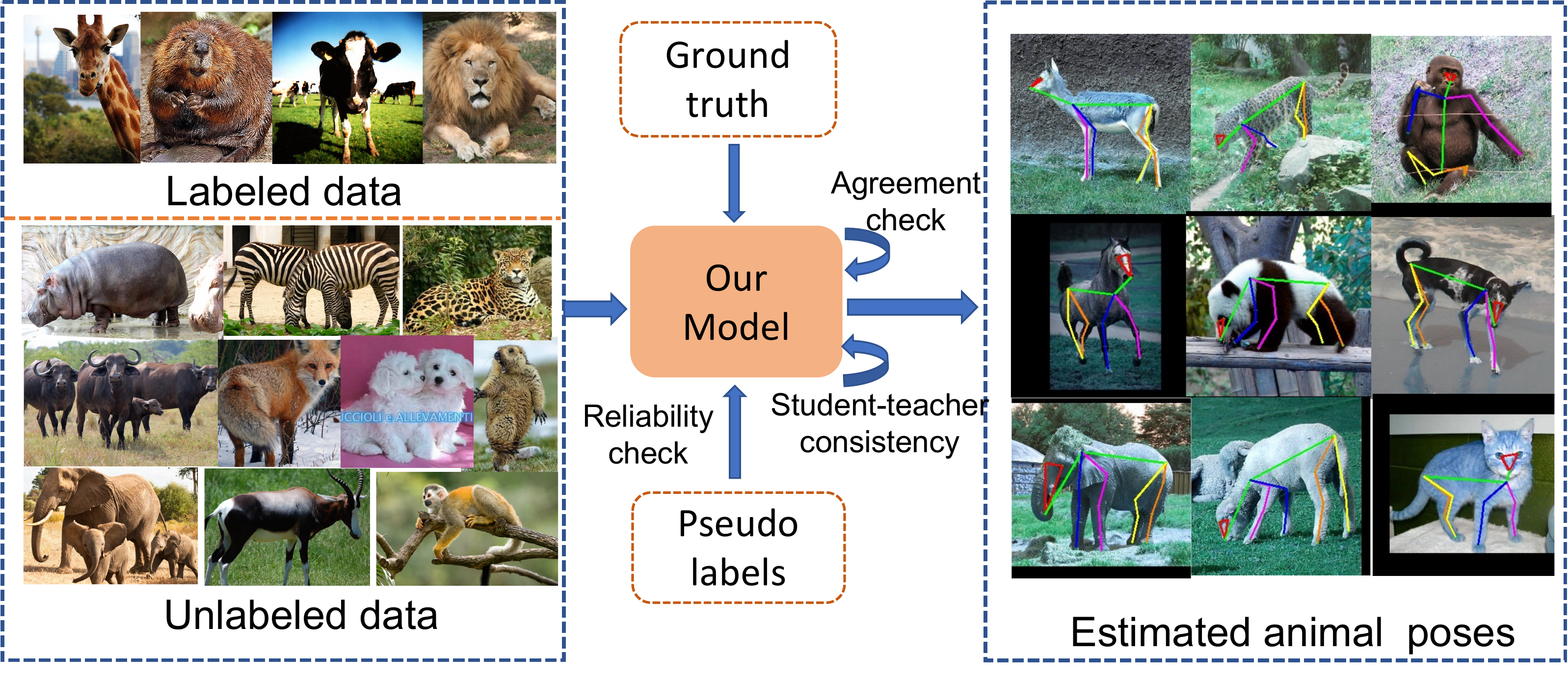}
    \caption{An illustration of our task. We aim to achieve accurate animal pose estimation with a small set of labeled images as well as unlabeled images.}
    \label{teaser}
\end{figure}

The ability to understand animal behavior is fundamental to many applications, such as farming, ecology and surveillance. Animal pose estimation 
is a key step to understand animal behavior 
and has attracted increasing attention in recent years \cite{cao2019cross, mu2020learning, li2021synthetic, shooter2021sydog, yu2021ap}. Although great success 
is achieved for human pose estimation with the development of sophisticated deep learning models, these techniques cannot be directly used for animals 
due to the lack of labeled animal pose data. 
Existing works overcome this problem by learning from human pose data \cite{cao2019cross} or synthetic animal images \cite{mu2020learning,li2021synthetic, shooter2021sydog}. However, there is a large domain gap between the real and synthetic (human) data. For example, the synthetic animal images in \cite{mu2020learning}, which are generated from CAD models, only exhibit limited pose, appearance and background variations. As a result, the model trained with the synthetic data may not adapt well to real images, especially for images with crowded scene or self-occlusion. Moreover, 
the generation of synthetic images is a tedious process 
that also requires expert knowledge. The above-mentioned problems lead us to ask whether 
we can achieve accurate animal pose estimation with minimal effort for annotating? To answer this question, we focus on how to learn from scarce labeled data for animal pose estimation. As shown in Fig.~\ref{teaser}, we aim to achieve accurate animal pose estimation with only a small set of labeled images 
and unlabeled images.

The data scarcity problem is solved by semi-supervised learning (SSL) for the classification task \cite{berthelot2019mixmatch,sohn2020fixmatch, zhang2021flexmatch, tarvainen2017mean}. One powerful class of SSL is pseudo labeling (PL), where artificial labels generated from a pretrained model are used together with labeled data to train a model. 
Impressive results have been achieved by applying PL \cite{cao2019cross, mu2020learning, li2021synthetic} to domain adaptation based animal pose estimation. However, the PL-based methods suffer from confirmation bias when the network starts to output incorrect predictions with high confidence, which we refer as noisy pseudo labels. To solve this noisy pseudo label problem, we use the small-loss trick \cite{han2018co, li2021synthetic} 
to select reliable pseudo labels, 
\ie~samples with small loss to update the network. This criteria is based on the memorization effect \cite{arpit2017closer} of deep networks 
where the network learn from clean and easy instances first before eventually overfitting to noisy ones. 
Consequently, clean samples exhibit smaller loss compared to noisy ones at the beginning of training. Although the small-loss trick is effective in selecting reliable pseudo labels, 
it is wasteful because numerous high-loss samples are left unused.

In this paper, we propose the ScarceNet, 
a PL-based 
approach to learn an animal pose network from scarce 
annotations. We first train an animal pose estimation network on the small set of labeled data, and then 
generate pseudo labels for the unlabeled images with the pretrained model. The pseudo labels are 
often corrupted with 
noise 
due to the small size of the training data. 
We thus also select reliable pseudo labels using the small-loss trick, However, instead of ignoring or adding consistency regularization to the high-loss samples as done in previous works \cite{mu2020learning, li2021synthetic}, we propose to identify reusable samples from those high-loss samples based on an agreement check. The agreement check is to guarantee that the model predictions for the reusable samples 
are smooth and concurrently far from the decision boundary, and hence 
allow us to 
directly re-label reusable samples with the model predictions. Finally, we introduce a student-teacher network \cite{tarvainen2017mean} and enforce a consistency constraint since there are still samples that are neither reliable nor reusable. By combining the reliable pseudo label selection with the reusable sample re-labeling and the consistency constraint, we 
can fully exploit the unlabeled data.  

In addition to the training scheme, we also modify the state-of-the-art human pose estimation network HRNet \cite{sun2019deep} and adapt it to the pseudo label-based learning framework. The original HRNet regresses the heatmaps of different joints from a shared feature representation. 
However, the negative effect of noisy pseudo label for any joint 
is propagated to all other joints through this shared feature representation. To mitigate this problem, we introduce a multi-branch HRNet (MBHRNet) where a joint-specific feature representation is learned for regressing the corresponding joint location in each branch.

We validate our approach on the challenging AP-10K dataset \cite{yu2021ap}, which includes animal images from more than 50 species. Our approach outperforms existing SSL methods by a large margin, especially when the labeled data 
is very limited (for example, 5 labeled images per species). We also test our approach under the inter-family transfer setting, where animal species from only one family are labeled and the rest are unlabeled. Our approach improves the HRNet baseline significantly, and even outperform the supervised counterpart for some species. We further test on the TigDog dataset \cite{mu2020learning}, where we can outperform existing synthetic data based approaches when only 0.5\% of the dataset is labeled. Our contribution can be summarized as follows:
\begin{itemize}
    \item We propose a pseudo label based training scheme for the task of animal pose estimation from scarce labels.
    \item We introduce the MBHRNet to mitigate negative effect of noisy joints to other joints in the same image.  
    \item We achieve state-of-the-art performance on several animal pose benchmarks under different settings.
\end{itemize}

\section{Related Work}
\paragraph{Animal Pose Estimation.} Animal pose estimation, including both 2D \cite{mu2020learning, li2021synthetic, cao2019cross} and 3D \cite{biggs2020left, li2021coarse, zuffi2019three}, has become an active research field in the last few years 
. The development of animal pose estimation techniques is relatively slower than the human counterpart due to the lack of large scale animal pose dataset like COCO \cite{lin2014microsoft} or MPII \cite{andriluka20142d}. Existing works circumvent the requirement for animal pose data by transferring knowledge from other domains, such as human pose data \cite{cao2019cross} and synthetic animal data \cite{mu2020learning, li2021synthetic, shooter2021sydog}. 
Although impressive results have been achieved in \cite{cao2019cross}, 
substantial amounts of labeled animal images are still needed 
to facilitate the knowledge transfer. This is due to the large domain gap between human and animal images, including pose, appearance and body structure differences. To reduce the domain gap between the source and target domains, synthetic animal images are generated from CAD models and used as the source images in \cite{mu2020learning}. Three consistency criteria are proposed to generate pseudo labels for real images. Improved upon \cite{mu2020learning}, Li \etal \cite{li2021synthetic} propose a pseudo label updating strategy to prevent the network from overfitting to the noisy pseudo labels. Despite the domain gap is reduced by using synthetic animal images, the limited pose and appearance variations of the CAD models prohibit the synthetic data based approaches from adapting to real scenarios with crowded scene or self-occlusion. Moreover, synthesize realistic images is a tedious process, which also requires expert knowledge. In view of this, we focus on learning from scarce annotations 
to minimize human labor and achieve accurate animal pose estimation at the same time. 

\paragraph{Semi-supervised Learning.}
Semi-supervised learning is powerful in leveraging unlabeled data to improve a model's performance when labeled data is limited. One of the most widely used SSL techniques is pseudo labeling \cite{lee2013pseudo,xie2020self}, which generates artificial labels for unlabeled images from model predictions. Another technique is consistency regularization \cite{rasmus2015semi,laine2016temporal,tarvainen2017mean}, which enforces that the model output should be consistent when the input is randomly perturbed. MixMatch \cite{berthelot2019mixmatch} combines the consistency regularization with the entropy minimization to encourage the network to output confident predictions for unlabeled data. More recent Unsupervised Data
Augmentation (UDA) \cite{xie2020unsupervised} and FixMatch \cite{sohn2020fixmatch} combine pseudo labeling, consistency regularization with strong augmentations \cite{cubuk2020randaugment} and achieve superior performance. 
FlexMatch \cite{zhang2021flexmatch} further improves over UDA and Fixmatch by introducing a curriculum learning scheme. 
We 
also adopt the PL-based techniques in this work given its effectiveness. To mitigate the negative effect of 
noisy pseudo labels, we combine 
PL with reliable pseudo label selection and reusable sample re-labeling.

\paragraph{Learning from Noisy Labels.} 
Learning from noisy labels has become an important task since deep neural networks are known to be susceptible to noisy labels. To reduce the negative effect of noisy labels, different approaches which can be divided into three categories have been proposed. The first category is to treat the true labels as hidden variables and use an extra network to estimate a noise transition matrix \cite{Goldberger2017TrainingDN, patrini2017making}. These methods assume there is correlation between the corrupted and clean labels, hence cannot handle complicate cases with random noise. The second category is to design noise-robust loss functions \cite{ghosh2017robust}, such as the generalized cross entropy \cite{zhang2018generalized} and the symmetric cross entropy \cite{wang2019symmetric}. The robust-loss based approaches are theoretically sound, but may only be capable of handling a certain noise rate. The last category \cite{jiang2018mentornet, malach2017decoupling, han2018co} focuses on selecting potential clean samples 
to update the network. 
One promising selection criteria is the small-loss trick \cite{han2018co}, 
which is designed based on the memorization effect of deep networks and has been widely applied to different tasks \cite{song2019selfie, yao2021jo, li2021synthetic}. We 
also apply the small-loss trick to select reliable pseudo labels. However, instead of ignoring the high-loss samples as done in previous works, we combine the sample selection with the reusable sample re-labeling to make full use of the unlabeled data.

\section{Our Method: ScarceNet}
\begin{figure*}
    \centering
    \includegraphics[width=0.998\linewidth]{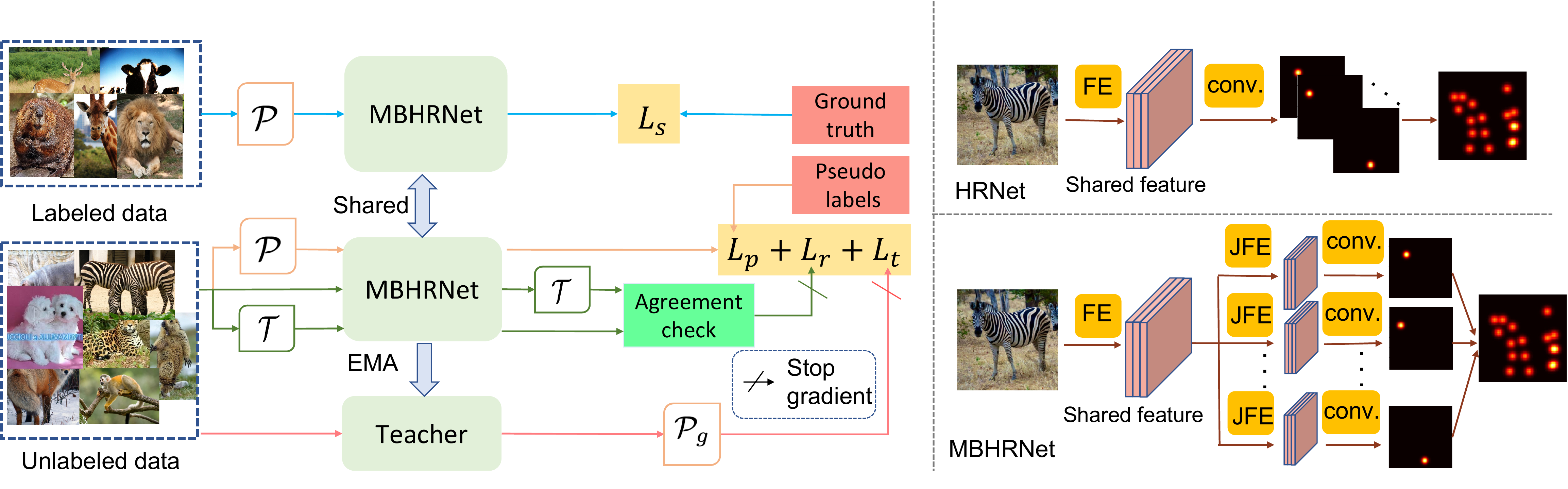}
    \caption{The left part shows the overall framework of our ScarceNet, where $\mathcal{P}$ and $\mathcal{T}$ represent strong and weak augmentation, respectively. The right part shows a comparison between the HRNet (top) and the MBHRNet (bottom), where `FE' represents feature extractor, `conv.' represents convolutional layer and `JFE' represents joint specific feature extractor.}
    \label{network}
\end{figure*}

We propose a PL-based approach to learn animal pose estimation from scarce annotations. The overall framework of our 
ScarceNet is illustrated in the left part of Fig.~\ref{network}, where we propose a multi-branch HRNet (MBHRNet) 
as our animal pose estimation network. Let us denote an input image as $\textbf{x}$ and its ground truth animal pose (if available) as $\textbf{y}$. Given a small set of labeled images
$\mathcal{D}_l = \{ \textbf{x}_i^l, \textbf{y}_i^l \}_{i=1}^{N_l}$ and unlabeled images 
$\mathcal{D}_u=\{ \textbf{x}_i^u\}_{i=1}^{N_u}$, the objective is to train the animal pose estimation network by leveraging both the labeled and unlabeled data. $N_l$ and $N_u$ represent the number of labeled and unlabeled images, respectively, and $N_l \ll N_u$. The labeled images are supervised with the ground truth pose data. 
We then generate pseudo labels with the model trained on the labeled data for the unlabeled images.
However, these generated pseudo labels are usually heavily corrupted with noise, especially when the labeled data is very limited. We thus apply the small-loss trick \cite{han2018co} to select a set of reliable pseudo labels. Despite its effectiveness, pseudo label selection by the small-loss trick tends to discard numerous high-loss samples. This results in high 
wastage since those discarded samples can still provide extra information for better discrimination. 
In view of this, we propose a reusable sample re-labeling step to further identify reusable samples from the high-loss samples via an agreement check and re-generate the corresponding pseudo labels for supervision. Lastly, we design a student-teacher framework to enforce consistency between the outputs of the student and teacher network. 

\subsection{Reliable Pseudo Label Selection}
We first train the animal pose estimation network 
on labeled images $\mathcal{D}_l$ with a supervised loss:
\begin{equation}
    \label{loss_sup}
    \mathcal{L}_\text{s} = \frac{1}{J} \sum_{i=1}^J \| H_i - \hat{H}_i \|^2,
\end{equation}
where $H_i$ is the ground truth heatmap, 
$J$ denotes the total number of joints, and $\hat{H}_i$ represents the output heatmap of the network. The pretrained network is then used to generate an initial set of pseudo labels for the unlabeled images
$\mathcal{D}_u$ following the procedure in \cite{mu2020learning}. The generated pseudo labels are generally noisy and 
can hurt the performance if directly used for training. To prevent this, we select reliable pseudo labels, denoted as $\mathbb{S}$, from the initial pseudo labels using the small-loss trick. Specifically, we only make use of samples with loss smaller than a threshold to update the network. 
Formally, this can be expressed as:
\begin{equation}
    \label{loss_pseudo}
    \mathcal{L}_\text{p} = \frac{\sum_{i=1}^J \{\mathcal{L}^i_\text{p} < l_\text{r}\}_1 \mathcal{L}^i_\text{p}}{\sum_{i=1}^J \{\mathcal{L}^i_\text{p} < l_\text{r}\}_1}, ~ \text{where}~ \mathcal{L}^i_\text{p} = \| \Tilde{H}_i - \hat{H}_i \|^2.
\end{equation}
$\tilde{H}_i$ represents the pseudo label for the $i^\text{th}$ joint and $\mathcal{L}^i_\text{p}$ is the corresponding pseudo label based loss. $\{\text{condition}\}_1$ is a condition function, which outputs 1 when the condition is true and 0 if false. $l_\text{r}$ denotes the threshold loss value, and more details on how to decide $l_\text{r}$ are provided in the supplementary material. 
This small-loss trick is designed based on the memorization effect of deep network, and has been shown effective in selecting reliable labels in previous works \cite{song2019selfie, li2021synthetic}.

\subsection{Reusable Sample Re-labeling}
Only small-loss samples are 
selected in the reliable pseudo label selection process, and the 
remaining pseudo labels with large loss are discarded. This is 
improvident since the high-loss samples 
can also contribute to improving the network if used properly. To this end, we propose to further identify reusable samples, denoted as $\mathbb{U}$, from the unused samples to fully exploit the unlabeled data. We define reusable samples as the ones that may also provide reliable knowledge to the network despite their large losses. To identify the reusable samples, we introduce an agreement check. Specifically, given an input $\textbf{x}^u$, we generate another view of this sample by adding a random transformation $\mathcal{T}$, which consists of rotation and scaling. Both samples are fed into the pose estimation network to produce the corresponding heatmaps, denoted as:
\begin{equation}
    H^{\text{v}_1} = f(\textbf{x}^u), H^{\text{v}_2} = f(\mathcal{T}\textbf{x}^u).
\end{equation}
The location for the $i^{th}$ joint is obtained by applying $\operatorname{argmax}$ operation to the corresponding output heatmap:
\begin{equation}
    \textbf{h}_i^{\text{v}_1} = \operatorname{argmax}(H_i^{\text{v}_1}), \textbf{h}_i^{\text{v}_2} = \operatorname{argmax}(H_i^{\text{v}_2}).
\end{equation}
Unlike the classification task where the output is transformation invariant \cite{ke2019dual, yao2021jo}, the output joint location is transformation 
equivariant. 
Consequently, we have to transform $ \textbf{h}_i^{\text{v}_1}$ 
before the agreement check, \ie: 
\begin{equation}
    d_i = \| \mathcal{T}\textbf{h}_i^{\text{v}_1} -\textbf{h}_i^{\text{v}_2}\|^2.
\end{equation}
Subsequently, we can determine whether a sample is reusable based on the distance between the outputs of the two views. 
Specifically, a sample $\textbf{x}^u_i$ is reusable 
when the distance fulfills $d_i < d_\text{r}$, 
where $d_\text{r}$ represents the distance threshold. This agreement check is based on the agreement maximization principle \cite{sindhwani2005co}, which has also been used for out-of-distribution detection \cite{yao2021jo} 
and stable sample selection \cite{ke2019dual}. Intuitively, the reusable samples must satisfy 
smoothness in the neighborhood of the model prediction 
and 
must also be far from the decision boundary. Otherwise, the output 
becomes inconsistent when the input is randomly transformed.

The next step is to re-label the identified reusable samples since they exhibit large loss based on the initial pseudo labels. We directly use the prediction from the current model as the new pseudo labels to train the network. The motivation is twofold: 1) The model has been trained with unlabeled images, hence it is able to output more accurate prediction comparing to the model trained only on the labeled images. 2) The selection criteria for reusable samples indicates that the current model prediction is already reliable. Finally, the loss function for the reusable samples can be defined as:
\begin{equation}
\label{correction_loss}
        \mathcal{L}_\text{r} = \frac{\sum_{i=1}^J \{d^i < d_\text{r}\}_1 \mathcal{L}^i_\text{r}}{\sum_{i=1}^J \{d^i < d_\text{r}\}_1}, ~ \text{where}~ \mathcal{L}^i_\text{r} = \| \bar{H}_i - \hat{H}_i \|^2.
\end{equation}
$\bar{H}_i$ represents the heatmap generated from the joint location of current model prediction. By selecting the reusable samples, we are able to make use of more unlabeled images and prevent the network from overfitting to the initial noisy pseudo labels at the same time.

\subsection{Student-teacher Consistency}
Both the reliable pseudo label selection and reusable sample selection are based on specific constraints. Samples that do not fulfill either constraint are still left unused. In view of this, we further introduce a teacher network and enforce the consistency between the student and teacher network. This is inspired by the mean teacher 
framework \cite{tarvainen2017mean}, which enforces the consistency constraint for unlabeled data for semi-supervised classification. The teacher network shares the same structure with the student network, and the weights are updated with an exponential moving average (EMA) of the student model:
\begin{equation}
\label{teacher_update}
    \theta^{t}_{\text{mt}} = \alpha  \theta^{t-1}_{\text{mt}} + (1 - \alpha) \theta^t.
\end{equation}
 $\theta_{\text{mt}}$ and $\theta$ denote the weights of the teacher and student network, respectively. The superscript $t$ represents the training step, and $\alpha$ is a smoothing coefficient of the EMA update. We can see that the teacher network is a temporal ensemble of networks, and thus it can provide more accurate pseudo labels. As shown in Fig.~\ref{network}, we add random perturbations $\mathcal{P}$ to the input of the student network, which consists of both geometry-based perturbations $\mathcal{P}_g$ such as rotation, scaling and flipping, and image-based perturbation $\mathcal{P}_m$ such as noise, occlusion and blurring. Consistency is then enforced between the teacher and student network 
 as:
 \begin{equation}
    \label{consistency_loss}
     \mathcal{L}_\text{t} =\frac{1}{J} \sum_{i=1}^J \|\mathcal{P}_g H_i^t - \hat{H}_i\|^2,
 \end{equation}
where $H_i^t$ denotes the output of the teacher network. Note that only the geometry-based perturbation $\mathcal{P}_g$ is added to the output of the teacher network since the output joint location is invariant to the image-based perturbation.

\subsection{Multi-head Animal Pose Estimation Network}
We adopt the human pose estimation network HRNet as our backbone. 
The HRNet has achieved state-of-the-art performance for human pose estimation by combining high-resolution feature representations with multi-resolution feature fusion.
However, as shown in the right top part of Fig.~\ref{network}, the final heatmaps of HRNet are regressed from a shared feature representation with one linear layer. This shared mechanism 
can have negative effect in the presence of noisy labels. For example, the feature representation 
is affected 
by backpropagation when the pseudo label for any joint is noisy 
and thus further affect the prediction of other joints in the same image. To mitigate this problem, we introduce a multi-branch head to the HRNet to learn a specific feature representation for each joint. As shown in the right bottom part of Fig.~\ref{network}, the network has $J$ output branches, where each branch learns a specific feature representation via a joint specific feature extractor (JFE) to regress the corresponding joint location. In this way, the adverse effect from the
noisy pseudo label of a joint gets absorbed in its respective
branch.

\subsection{Strong-weak Augmentation}
We also introduce a strong-weak augmentation training scheme, where two sets of augmentation are applied to the inputs, 
denoted as $\mathcal{T}$ and $\mathcal{P}$ in previous sections. 
Particularly, we apply strong augmentations $\mathcal{P}$ for the learning task with backpropagation and weak augmentation $\mathcal{T}$ for the re-labeling task.
As shown in Fig.~\ref{network}, strong augmentation $\mathcal{P}$ is applied to the labeled images in the blue path and unlabeled images in the orange path. These two paths are responsible for updating the network, and hence strong augmentation 
can improve the generalization capacity. In comparison, the weak augmentation $\mathcal{T}$ is applied to the green path because this path is responsible for identifying and re-labeling reusable samples. Overly-strong augmentation 
can have adversarial affects on the agreement check process since strong augmentation tends to result in inconsistent predictions. More details on the augmentation are provided 
in the supplementary material.

\subsection{Overall Framework}
We train our network in two stages. In the first stage, we train the MBHRNet on $\mathcal{D}_l$ with the supervised loss in Eqn.~\eqref{loss_sup}. The trained network is applied to generate pseudo labels for the unlabeled data. In the second stage, the student network is initialized with the pretrained model and trained with both $\mathcal{D}_l$ and $\mathcal{D}_u$. The objective function for the second stage consists of both labeled and unlabeled losses as follows: 
\begin{equation}
\label{total_loss}
    \mathcal{L} = \lambda_1 \mathcal{L}_\text{s} + \lambda_2 \mathcal{L}_\text{p} + \lambda_3 \mathcal{L}_\text{r} + \lambda_4 \mathcal{L}_\text{t},
\end{equation}
where $\lambda_1$,  $\lambda_2$,  $\lambda_3$, and $\lambda_4$ are the weights for different losses. The overall training procedure is presented in Algorithm \ref{train_procedure}.
\begin{algorithm}[h]
\caption{Overall training procedure of the second stage}
\label{train_procedure}
\textbf{Input}: labeled data $\mathcal{D}_l$ and unlabeled data $ \mathcal{D}_u$, 
threshold loss value $l_r$,
distance threshold $d_r$, 
batch size $B$
\begin{algorithmic}[1]
\FOR{$\{x_1, x_2, ..., x_B\} \in \mathcal{D}_l \cup \mathcal{D}_u$}
\IF{$x_i \in \mathcal{D}_l$}
\STATE Compute supervised loss $\mathcal{L}_\text{s}$ according to Eqn.~\eqref{loss_sup}
\ELSE 
\IF{$x_i \in \mathbb{S}$ }
\STATE Compute the pseudo labeled based loss $\mathcal{L}_\text{p}$ according to Eqn.~\eqref{loss_pseudo}
\ELSIF{$x_i \in \mathbb{U}$}
\STATE Compute loss $\mathcal{L}_\text{r}$ for reusable samples by Eqn.~\eqref{correction_loss}
\ENDIF
\STATE Compute the consistency loss $\mathcal{L}_\text{t}$ according to Eqn.~\eqref{consistency_loss}
\ENDIF
\STATE Compute the total loss $\mathcal{L}$ according to Eqn.~\eqref{total_loss}
\STATE Update student network $\theta \leftarrow \theta - \eta \Delta \mathcal{L}$
\STATE Update teacher network by Eqn.~\eqref{teacher_update}
\ENDFOR
\end{algorithmic}
\end{algorithm}
\vspace{1mm}
\section{Experimental Results}

\paragraph{Training details.} We adopt the HRNet-w32 as the backbone and train our network in two stages. We first train the MBHRNet on the labeled data with the supervised loss. 
Following \cite{sun2019deep}, we train the network with the Adam optimizer for 210 epochs. The initial learning rate is set to $10^{-3}$, and is dropped to $10^{-4}$ and $10^{-5}$ at the $170^{th}$ and $200^{th}$ epochs, respectively.
 The trained model is then used to generate pseudo labels following \cite{mu2020learning}. In the second stage, we initialized the student network from the first stage and train the whole framework with both labeled and unlabeled data for another 150 epochs. More training details are provided in the supplementary material.

\paragraph{Datasets.} We evaluate our method on the AP-10K dataset \cite{yu2021ap} and the TigDog dataset \cite{mu2020learning}. 
The AP-10K dataset is a recently proposed animal pose dataset, which contains more diverse animal species than previous datasets
\cite{mu2020learning, cao2019cross}.
The dataset contains 10,015 labeled images in total, which are split into train, validation, and test sets with a ratio of 7:1:2 per animal species. 
The TigDog dataset includes 8380 images for the horse category and 6523 images for the tiger category. This dataset is used in several domain adaptation works \cite{mu2020learning,li2021synthetic} to show that the knowledge learned from synthetic animals can be transferred to real animal images. We also evaluate on this dataset to show that better performance can be achieved by labeling just a few images, hence can reduce human labor compared to generating synthetic data.

\paragraph{Evaluation metrics.} We adopt the Mean Average Precision (mAP) as the evaluation metric for the AP-10K dataset.
The Percentage of Correct Keypoints (PCK), which computes the percentage of joints that are within a normalized distance to the ground truth locations, is used for the TigDog dataset
following \cite{mu2020learning}.

\begin{table}[h!]
\centering
\setlength{\tabcolsep}{6pt}
\begin{tabular*}{0.48\textwidth}{c|c c c c c}
\hline
    & 5 & 10 & 15 & 20 & 25  \\ \hline
  HRNet \cite{sun2019deep} & 0.360 & 0.463 & 0.511 & 0.547 & 0.588 \\ \hdashline
  UDA \cite{xie2020unsupervised}  & 0.429 & 0.519 & 0.566 & 0.580 & 0.628 \\
  FixMatch \cite{sohn2020fixmatch} & 0.478 & 0.544 & 0.589 & 0.601 & 0.631 \\
  FlexMatch \cite{zhang2021flexmatch} & 0.466 & 0.555 & 0.596 & 0.618 & 0.646 \\
  Ours   & \textbf{0.533} & \textbf{0.597} & \textbf{0.632} & \textbf{0.654} & \textbf{0.681} \\ \hline
  
\end{tabular*}
\caption{Results (mAP) on the AP-10K when 5, 10, 15, 20 and 25 images per species are labeled. Best results in bold.}
\label{ap10k-results1}
\end{table}

\begin{table*}[ht!]
\centering
\setlength{\tabcolsep}{10pt}
\begin{tabular*}{0.85\textwidth}{c c c | c c c}
\hline
  Species  & Setting & Performance & Species & Setting & Performance  \\ \hline
 \multirow{4}{*}{Deer} & Generalization & 0.723 & \multirow{4}{*}{Moose} & Generalization & 0.587\\
                        & Few-Shot & 0.742 &                         & Few-Shot & 0.648  \\
                        & Transfer & 0.751 &                        & Transfer & 0.726 \\
                        & Ours & 0.767 &                        & Ours & 0.547  \\ \hline
 \multirow{4}{*} {Horse}  & Generalization & 0.592 & \multirow{4}{*} {Zebra} & Generalization & 0.324\\
                        & Few-Shot & 0.635 &                     & Few-Shot & 0.480  \\
                        & Transfer & 0.718 &                    & Transfer & 0.708   \\
                        & Ours & 0.749 &                        & Ours & 0.562 \\ \hline
 \multirow{4}{*} {Chimpanzee} & Generalization & 0.009 & \multirow{4}{*} {Gorilla} & Generalization & 0.017 \\                         
                        & Few-Shot & 0.022 &                     & Few-Shot & 0.144  \\
                        & Transfer & 0.550 &                    & Transfer & 0.662  \\
                        & Ours & 0.057 &                        & Ours & 0.033  \\ \hline
  
\end{tabular*}
\caption{Results (mAP) for the Cervidae, Equidae and Hominidae
when only images from the Bovidae are labeled.}
\label{ap10k-results2}
\end{table*}

\subsection{Results on the AP-10K Dataset.}

We evaluate our approach under two scenarios on the AP-10K dataset. In the first scenario, a set of images are labeled for each animal species. The second scenario is that only the images in one family are labeled and we test on other animal families. This setting is more challenge since the appearance, size and background environment are quite different for different animal species.

\begin{figure*}[h!]
    \centering
    

    \includegraphics[width=0.135\textwidth]{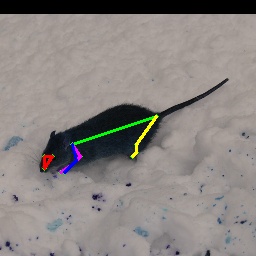}
  \includegraphics[width=0.135\textwidth]{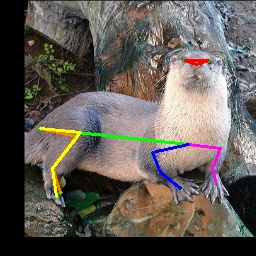}
    \includegraphics[width=0.135\textwidth]{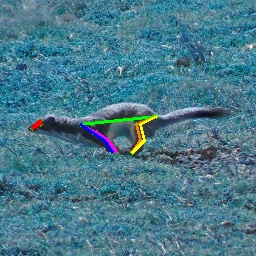}
     \includegraphics[width=0.135\textwidth]{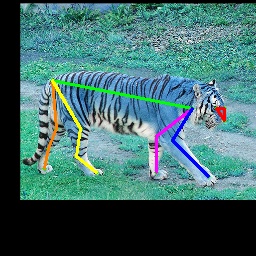}
    \includegraphics[width=0.135\textwidth]{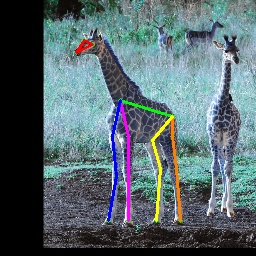}
    \includegraphics[width=0.135\textwidth]{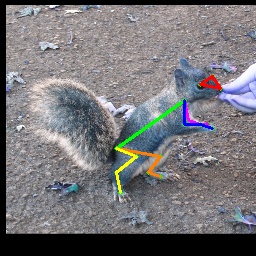} \\

    \includegraphics[width=0.135\textwidth]{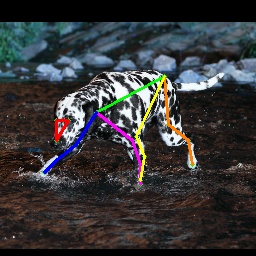}
    \includegraphics[width=0.135\textwidth]{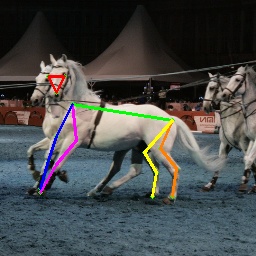}
   \includegraphics[width=0.135\textwidth]{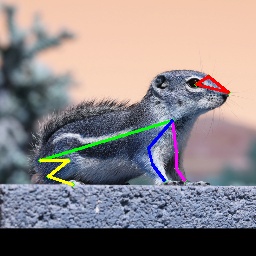}
    \includegraphics[width=0.135\textwidth]{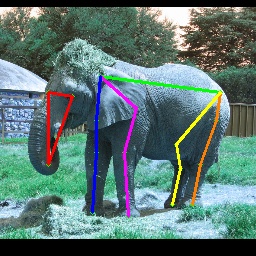}
    \includegraphics[width=0.135\textwidth]{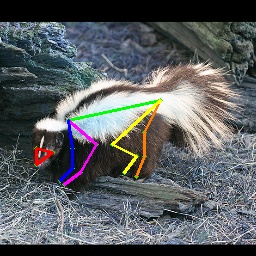}
    \includegraphics[width=0.135\textwidth]{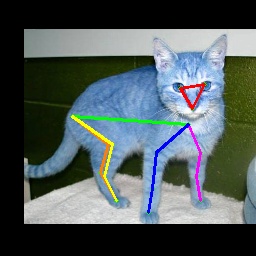} \\

  \includegraphics[width=0.135\textwidth]{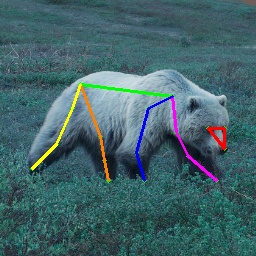}
    \includegraphics[width=0.135\textwidth]{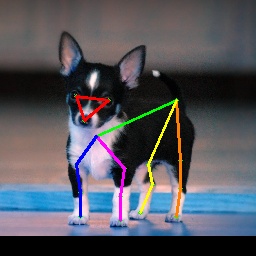}
    \includegraphics[width=0.135\textwidth]{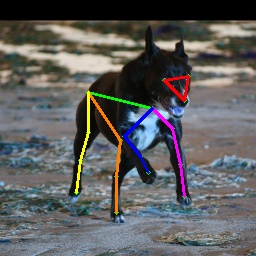} 
    \includegraphics[width=0.135\textwidth]{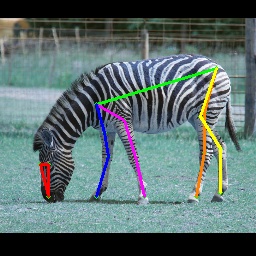}
    \includegraphics[width=0.135\textwidth]{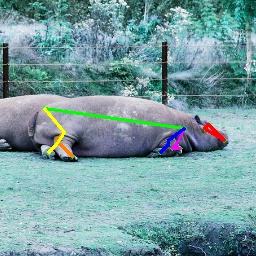}
    \includegraphics[width=0.135\textwidth]{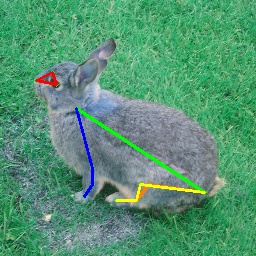} \\
    
    \includegraphics[width=0.135\textwidth]{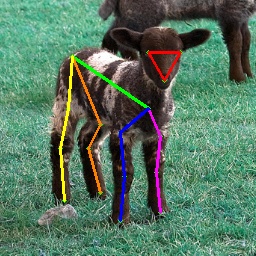} 
    \includegraphics[width=0.135\textwidth]{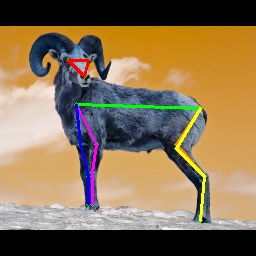} 
    \includegraphics[width=0.135\textwidth]{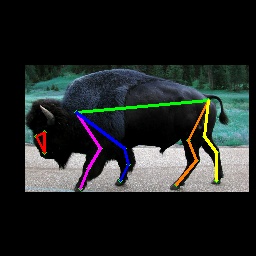} 
    \includegraphics[width=0.135\textwidth]{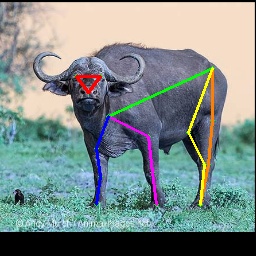} 
    \includegraphics[width=0.135\textwidth]{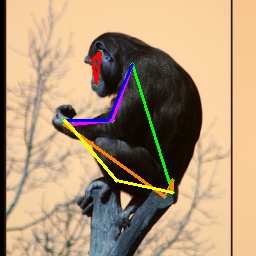} 
    \includegraphics[width=0.135\textwidth]{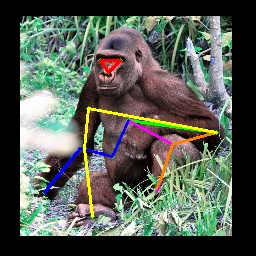} 
    \\
        \caption{Qualitative results for the AP-10K dataset when 25 images per species is labeled (except for the last two examples). The last two examples show the \textbf{failure cases} when only images from the Bovidae family are labeled. 
    } \vspace{2mm}
    \label{fig:qualitative}
\end{figure*}

We show results for the first scenario in Tab.~\ref{ap10k-results1}, where we evaluate when 5, 10, 15, 20 and 25 images per animal species are labeled. There are 50 species in total, which results in 250, 500, 720, 1000 and 1250 labeled images out of 7023 images. We compare with state-of-the-art semi-supervised approaches, including UDA \cite{xie2020unsupervised}, FixMatch \cite{sohn2020fixmatch} and FlexMatch\cite{zhang2021flexmatch}. `HRNet' represents HRNet trained only with labeled images. We reimplement those semi-supervised approaches according to the open repository since they only show results for classification task. We can see that our approach outperform existing approaches by a large margin, especially when very few labeled images are available, such as the case of 5 labels per species. The reason is that UDA, FixMatch and FlexMatch only retain the samples with reliable pseudo labels based on the confidence scores. In comparison, we combine the reliable pseudo label selection with the reusable sample re-labeling and teacher-student consistency, which helps to fully 
utilize the unlabeled data. Moreover, our results are the closest to the fully supervised setting \cite{yu2021ap}, \ie~0.738 mAP, when only 25 images per animal species are labeled. We also show qualitative results for the 25 labels per animal species setting in Fig.~\ref{fig:qualitative}. 

We also show results for the case when images from only one animal family are labeled in Tab.~\ref{ap10k-results2}. Following the inter-family transfer learning setting in \cite{yu2021ap}, we train our model with labeled images of the Bovidae family and the remaining unlabeled images, and test on families including Cervidae, Equidae and Hominidae.
There are seven animal species in the Bovidae family 
that includes 
a total of 755 images.  The results under the `Generalization', `Few-Shot' and `Transfer' settings are taken from \cite{yu2021ap}. `Generalization' means that the model is trained only with images in the Bovidae, `Few-Shot' means that 20 labels from the testing species are used, and `Transfer' means that all labels of the testing species are used. We can see that we achieve better performance than the generalization setting in all cases except for `Moose'. Moreover, we even outperform the `Transfer' setting for `Deer' and `Horse' although we do not use any labels from the target species.  
Our results for the `Chimpanzee' and `Gorilla' are not satisfactory although still comparable to the `Few-Shot' setting. We show the failure cases of `Chimpanzee' and `Gorilla' in the last two examples of Fig.~\ref{fig:qualitative}. We can see that `Chimpanzee' and `Gorilla' are very different from animal species in the Bovidae family (the first four examples of the last row) in terms of appearance and pose, which makes the knowledge transfer hard without using any labels from the target species.

\begin{table*}[ht!]
\centering
\small
\setlength{\tabcolsep}{8pt}
\begin{tabular*}{0.9\textwidth}{c|c:c c c c c c}
\hline
& Real & Cycgan \cite{zhu2017unpaired} & BDL \cite{li2019bidirectional}  & Cycada \cite{hoffman2018cycada}  & CC-SSL \cite{mu2020learning}  & MDAM-MT \cite{li2021synthetic} & Ours \\ \hline
Horse & 78.98 & 51.86 & 62.33 & 55.57 & 70.77 &  \textbf{79.50} & 73.05\\
Tiger & 81.99 & 46.47 & 52.26 & 51.48 & 64.14 & 67.76 & \textbf{74.88} \\
Average & 80.48 & 49.17 & 57.30 & 53.53 & 67.52 & 73.66 & \textbf{73.83}\\ \hline
\end{tabular*}
\caption{PCK@0.05 accuracy of the TigDog dataset. `Real'
means full supervision.
Cycgan, BDL, Cycada, CC-SSL, MDAM-MT are domain adaptation approaches using synthetic data. Our approach is trained with only 36 labeled images per category.}
\label{resuts-tigdog}
\end{table*}

\begin{table*}[ht!]
\centering
\small
\setlength{\tabcolsep}{1.8pt}
\begin{tabular*}{1.0\textwidth}{c|c c c c c c c c|c c c c c c c c|c}
\hline
& \multicolumn{8}{c|}{Horse Accuracy} & \multicolumn{8}{c|}{Tiger Accuracy} & \multirow{2}{*}{Average}\\
 & Eye & Chin & Shoulder & Hip & Elbow & Knee & Hooves & Mean 
& Eye &Chin & Shoulder & Hip & Elbow & Knee & Hooves & Mean \\ \hline

CC-SSL & 84.60 & 90.26 & 69.69 & \textbf{85.89} & 68.58 & 68.73 & 61.33 & 70.77
        & 96.75 & 90.46 & 44.84 & 77.61 & \textbf{55.82} & 42.85 & 64.55 & 64.14 & 67.52 \\ 
MDAM-MT & \textbf{91.05} & 93.37 & \textbf{77.35} & 80.67 & 73.63 & \textbf{81.83} & 73.67 & \textbf{79.50}
            & 97.01 & 91.18 & 46.63 & \textbf{78.08} & 50.86 & 61.54 & 70.84 & 67.76 & 73.66\\ 
Ours & 89.42 & \textbf{93.60} & 74.63 & 83.61 & \textbf{74.31} & 79.91 & \textbf{73.99} & 79.18 & \textbf{99.13} & \textbf{92.82} & \textbf{46.96} & 75.36 & 50.46 & \textbf{63.51} & \textbf{74.82} & \textbf{68.98} & \textbf{74.13} \\\hline
\end{tabular*}
\caption{PCK@0.05 accuracy of the TigDog dataset for domain adaptation setting. `Mean' represents the accuracy for each animal category, and `Average' represents the average accuracy for all animal categories. Best results in bold.}
\label{resuts-tigdog2}
\end{table*}

\subsection{Results on the TigDog Dataset}

\begin{table}[h!]
\centering
\setlength{\tabcolsep}{8pt}
\begin{tabular*}{0.47\textwidth}{c|c c c c c}
\hline
      & 5 & 10 & 15 & 20 & 25  \\ \hline
  Full  & 0.533 & 0.597 & 0.632 & 0.655 & 0.681 \\ \hdashline
  - RSR  & 0.521 & 0.586 & 0.621 & 0.633 & 0.668 \\
    - MT  & 0.487 & 0.568 & 0.614 & 0.628 & 0.659 \\
  - MB & 0.520 & 0.585 & 0.624 & 0.645 & 0.670 \\
 - AUG & 0.487 & 0.565 & 0.608 & 0.625 & 0.654 \\ \hline

\end{tabular*}
\caption{Ablation study on the AP-10K dataset when 5, 10, 15, 20 and 25 images per animal species are labeled.}
\label{ablation-studies}
\end{table}

The TigDog Dataset is used in previous domain adaptation works \cite{mu2020learning, li2021synthetic} to show that models trained from synthetic images can be used for real ones. We also evaluate on this dataset to show that our approach can achieve comparable or even better performance compared with state-of-the-art domain adaptation works with very few labeled images. 
Note that we use the same ResNet backbone as \cite{mu2020learning} for all experiments on the TigDog dataset for fair comparison. The results are shown in Tab.~\ref{resuts-tigdog}, where the numbers are directly taken from \cite{li2021synthetic}. `Real' denotes fully-supervised setting, Cycgan \cite{zhu2017unpaired}, BDL \cite{li2019bidirectional}, Cycada \cite{hoffman2018cycada}, CC-SSL \cite{mu2020learning}, MDAM-MT \cite{li2021synthetic} are domain adaptation approaches trained with synthetic animal images and ``Ours'' means our approach trained with 36 images per animal category. 
Note that 36 is the minimum labeled data needed for each category to achieve better performance than existing unsupervised domain adaptation approaches, \ie we only need to label 72 out of 14,903 images for the Tigdog dataset. 
The results demonstrate the advantage of our setting since annotating such few images
does not take much human labor. 

We also test our approach on the domain adaptation setting to further verify the effectiveness. The results are shown in Tab.~\ref{resuts-tigdog2}, which includes accuracy for different joints, the mean accuracy for each animal category and the average accuracy over all categories.
We can see that our approach achieves the best performance in most of the cases.  Note that our approach and CC-SSL have simpler networks compared to MDAM-MT, which introduce an extra refinement module for self-knowledge distillation. 
More training details are provided in the supplementary material.

\subsection{Ablation Study} 
We conduct ablation study on the AP-10K dataset. Each component is removed from the full model to verify the corresponding contribution. The results are shown in Tab.~\ref{ablation-studies}, where `RSR' denotes reusable sample re-labeling, `MT' denotes student teacher consistency, `MB' denotes multi-branch architecture and `AUG' denotes strong-weak augmentation. Note that we remove the weak augmentation
$\mathcal{T}$
and replace the strong augmentation
$\mathcal{P}$ with commonly used rotation and scale augmentation for the ablation of `AUG'.
We can see that the performance drops when each component is removed 
. Specially, the accuracy drops significantly for the 5 labels per animal species setting when the student teacher consistency or strong-weak augmentation is removed. This can be attributed to two reasons. Firstly, the strong-weak augmentation enforces the output of the strong augmented images to be consistent with the pseudo labels that are generated using weak augmented images. In this way, the network is enforced to learn a smooth representation 
around each image
, which helps to improve the performance especially in the low-data regime. Secondly, The initial 
set of 
pseudo labels are extremely noisy when the model is trained with very few labeled data. The teacher network, which is a temporal ensemble of network, is able to provide more stable supervision for the student network. 

\section{Limitations}
Our approach achieves impressive performance even when very few labeled data is available. However, there are still many remaining challenges for the task of animal pose estimation for limited labeled data.
The large variations of appearance and pose across different animal species makes it hard for the network to generalize well to different animal species, \eg train with labels from Bovidae and test on Hominidae. We leave this for future research. 

\section{Conclusion}
We propose a pseudo label based training scheme for animal pose estimation from scarce annotations. We combine the reliable pseudo label selection with the reusable sample re-labeling and the student-teacher consistency to make full use of the unlabeled data. We also introduce the MBHRNet to mitigate the negative effect of noisy joints. Extensive experiments have been conducted to verify the effectiveness of our proposed approach. 

\paragraph{Acknowledgement.}
This research is supported by the National Research Foundation, Singapore under its AI Singapore Programme (AISG Award
No: AISG2-RP-2021-024), and the Tier 2 grant MOE-T2EP20120-0011 from the
Singapore Ministry of Education.


\pagebreak


	\twocolumn[{%
		\renewcommand\twocolumn[1][]{#1}%
		\vskip .5in
		\begin{center}
			\textbf{\Large Supplementary Material for}\\
			\vspace*{4pt}
			\textbf{\Large ScarceNet: Animal Pose Estimation with Scarce Annotations} \\
			\vspace*{10pt}
			{\large
				Chen Li \quad \quad Gim Hee Lee\\
			}
			\vskip .5em
			{\large Department of Computer Science, National University of Singapore\\}
			{\tt\small lichen@u.nus.edu \quad \quad gimhee.lee@comp.nus.edu.sg}
			\vspace*{10pt}
		\end{center}
	}]

\setcounter{equation}{0}
\setcounter{figure}{0}
\setcounter{table}{0}
\setcounter{section}{0}
\setcounter{page}{9}

\paragraph{Details on how to decide the threshold loss value.}
We select the reliable pseudo labels based on the pseudo label based loss (Eqn. (2) in the main paper), 
\ie~only samples with loss smaller than the threshold $l_r$ 
are be used.  Note that it is difficult to directly set $l_\text{r}$ since $\mathcal{L}^i_\text{p}$ does not fall in a specific range, 
and therefore we decide $l_\text{r}$ based on a percentile score. Specifically, given a batch of images as input, we first compute the pseudo label based loss for all joints $ \mathcal{L}_\text{p}^B = \{\mathcal{L}^i_\text{p} |_{i=1}^{B \times J}\}$, where $B$ denotes the batch size. The threshold loss value is then computed as:
\begin{equation}
    \label{threshold_value}
    l_\text{r} = \operatorname{percentile}(\mathcal{L}_\text{p}^B, c),
\end{equation}
where $\operatorname{percentile}(\cdot,\cdot)$ returns the value of the $c^{th}$ percentile.

\paragraph{Strong-weak augmentation.} We apply two sets of augmentations in our framework as described in our main paper. Specifically, we leverage 
RandAugment \cite{cubuk2020randaugment} as the strong augmentation $\mathcal{P}$, where we remove the translation and shearing augmentation since the object needs to be at the center of the image for pose estimation. For the weak augmentation $\mathcal{T}$, we apply the widely used augmentation operations in human pose estimation, including rotation ($-20^\circ-20^\circ$) and scaling ($0.9-1.1$).

\paragraph{Network details.} We adopt the HRNet-w32 as our backbone with an input size of $256 \times 256 \times 3$. The HRNet maintains high-resolution representation through the whole network in order to learn more accurate features for pose estimation. This high-resolution feature is fed into a linear prediction layer to regress joint locations after fusing with features at lower resolutions. In our framework, we replace the single-branch prediction layer with a multi-branch prediction layer, where each branch consists of a Bottleneck residual block \cite{he2016deep} followed by a linear layer. The output channel size of the last linear layer is one because we only regress the heatmap for one joint in each branch. 

\paragraph{Training details.} We implement our network in Pytorch and the parameters are optimized using the Adam \cite{kingma2014adam} optimizer with the default parameters. We first train the MBHRNet on the labeled data with the supervised loss. Following \cite{sun2019deep}, we train the network with the Adam optimizer for 210 epochs. The initial learning rate is set to $10^{-3}$, and is dropped to $10^{-4}$ and $10^{-5}$ at the $170^{th}$ and $200^{th}$ epochs, respectively.
We then generate pseudo labels with the pretrained model following \cite{mu2020learning}. We remove samples with low confidence score, and apply the small-loss trick to select reliable pseudo labels from the remaining ones. The distance threshold in the agreement check is empirically set to 0.6 in our experiments. 
The weights for different loss terms are also set empirically as $\lambda_1 = 2$, $\lambda_2 = 1$, $\lambda_3 = 1$ and $\lambda_4 = 2$. 
The training of the first stage takes 2-5 hours depends on the number of the labeled data. The second stage takes around 21 hours on two RTX 2080Ti GPUs.

\paragraph{Training details for the domain adaptation setting.} We test the effectiveness of our approach in the domain adaptation setting in the main paper. We replace the MBHRNet in our framework with the ResNet backbone \cite{mu2020learning} for fair comparison. A domain discriminator \cite{li2021synthetic} is also applied in the feature space to facilitate the knowledge transfer from synthetic to real images. Note that our approach and CC-SSL \cite{mu2020learning} has simpler network compared to MDAM-MT \cite{li2021synthetic}, which introduces an extra refinement module for self-knowledge distillation.

\begin{figure*}[h!]
    \centering

    \includegraphics[width=0.15\textwidth]{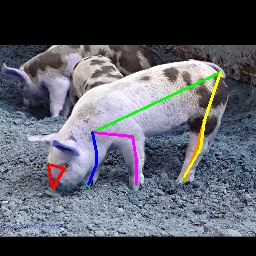}
    \includegraphics[width=0.15\textwidth]{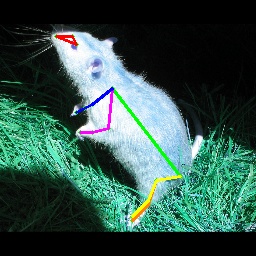}
    \includegraphics[width=0.15\textwidth]{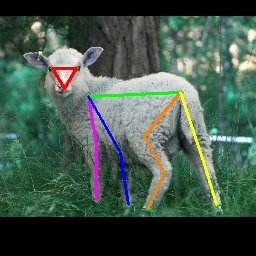}
    \includegraphics[width=0.15\textwidth]{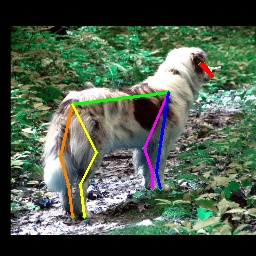}
  \includegraphics[width=0.15\textwidth]{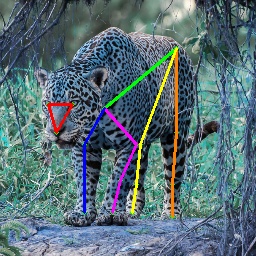}
    \includegraphics[width=0.15\textwidth]{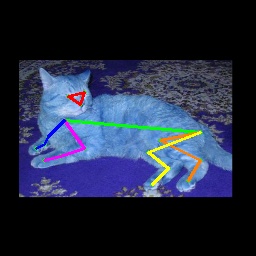} \\

    \includegraphics[width=0.15\textwidth]{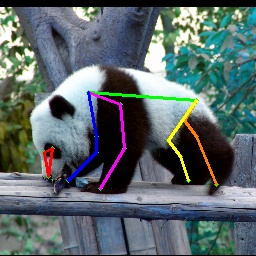}
    \includegraphics[width=0.15\textwidth]{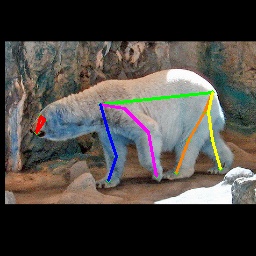}
    \includegraphics[width=0.15\textwidth]{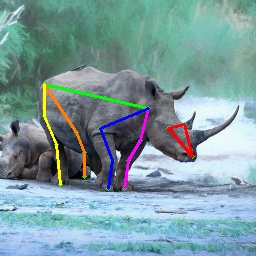}
    \includegraphics[width=0.15\textwidth]{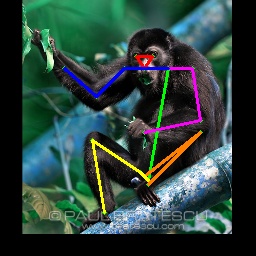}
     \includegraphics[width=0.15\textwidth]{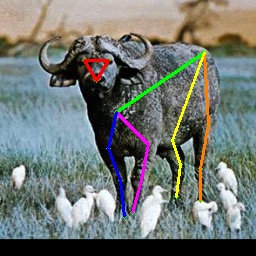}
    \includegraphics[width=0.15\textwidth]{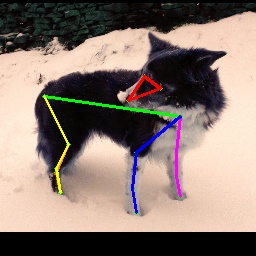} \\

     \includegraphics[width=0.15\textwidth]{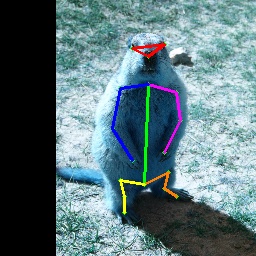} 
    \includegraphics[width=0.15\textwidth]{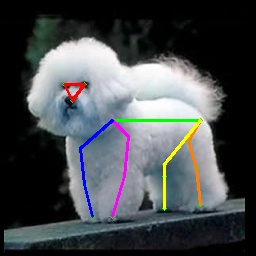}
    \includegraphics[width=0.15\textwidth]{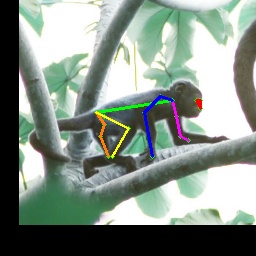}
    \includegraphics[width=0.15\textwidth]{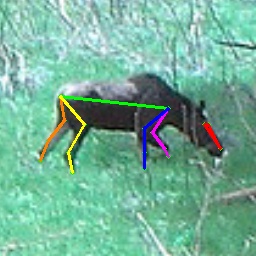}
  \includegraphics[width=0.15\textwidth]{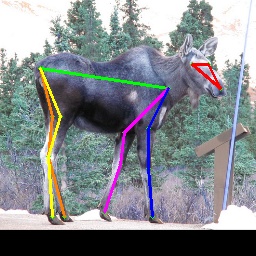}
    \includegraphics[width=0.15\textwidth]{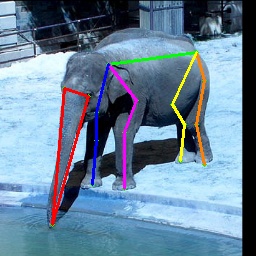} \\

    \includegraphics[width=0.15\textwidth]{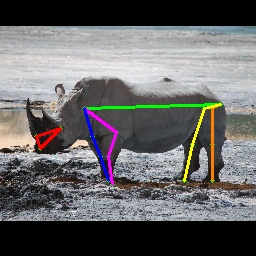}
    \includegraphics[width=0.15\textwidth]{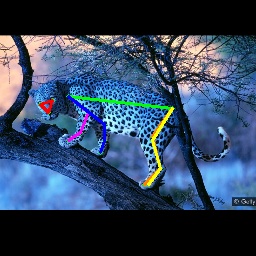}
    \includegraphics[width=0.15\textwidth]{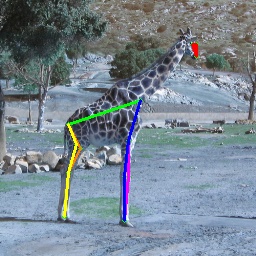}
    \includegraphics[width=0.15\textwidth]{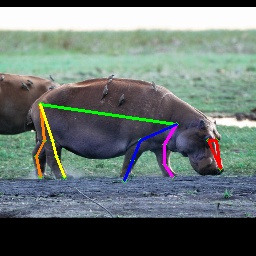}
  \includegraphics[width=0.15\textwidth]{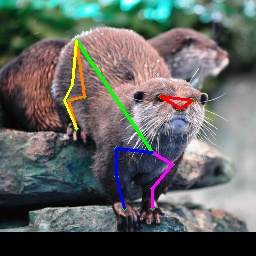}
    \includegraphics[width=0.15\textwidth]{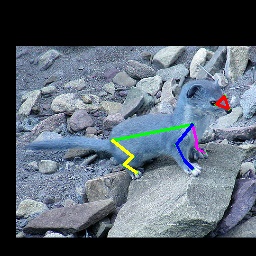} \\

        \caption{Qualitative results when 25 images per species are labeled.}
    \label{fig:qualitative_supp}
\end{figure*}

\paragraph{More qualitative results.} We show qualitative results for more animal species under the 25 labels per animal species setting in the first three rows of Fig.~\ref{fig:qualitative_supp}. We can see we that our network is able to estimate the pose accurately for diverse animal species when only 25 images for each species are labeled. 
\vspace{30mm}

{\small
\bibliographystyle{ieee_fullname}
\bibliography{egbib}
}

\end{document}